\documentclass[letterpaper, 10 pt, conference]{ieeeconf}  %
\IEEEoverridecommandlockouts                              %
\usepackage{fourier} 
\usepackage{array}
\usepackage{makecell}
\usepackage{dblfloatfix}
\usepackage{bm}
\usepackage{mathrsfs}

\usepackage{multibib}
\usepackage{multirow}
\usepackage{pifont}

\usepackage{lipsum}

\usepackage{subcaption}
\usepackage{amssymb,amsmath}

\usepackage{graphicx}
\usepackage{url}
\usepackage{cite} %
\usepackage{booktabs}
\usepackage{tabularx}

\usepackage[usenames,dvipsnames]{color}
\usepackage[x11names]{xcolor}

\definecolor{dark-gray}{gray}{0.30}
\definecolor{orange}{rgb}{1.0,0.5,0}

\usepackage{amsmath,amsfonts,bm}

\def\eqref#1{equation~\ref{#1}}

\def\1{\bm{1}}

\def\vtheta{{\bm{\theta}}}

\DeclareMathAlphabet{\mathsfit}{\encodingdefault}{\sfdefault}{m}{sl}
\SetMathAlphabet{\mathsfit}{bold}{\encodingdefault}{\sfdefault}{bx}{n}

\usepackage[shortcuts]{extdash}  %

\newcommand \cmame {CMA\=/ME}

\newcommand \mapelites {MAP\=/Elites}

\usepackage{xcolor}

\newcommand{\xxnote}[3]{}
\ifx\hidenotes\undefined
  \renewcommand{\xxnote}[3]{\color{#2}{(#1: #3)}}
\fi

\definecolor{lightgrey}{rgb}{0.9, 0.9, 0.9}

\newcommand{\fref}[1]{Fig.~\ref{#1}}

\usepackage{booktabs}
\usepackage{multirow}
\usepackage{array}
\newcolumntype{L}[1]
  {>{\raggedright\let\newline\\\arraybackslash\hspace{0pt}}m{#1}}
\newcolumntype{C}[1]
  {>{\centering\let\newline\\\arraybackslash\hspace{0pt}}m{#1}}
\newcolumntype{R}[1]
  {>{\raggedleft\let\newline\\\arraybackslash\hspace{0pt}}m{#1}}

\title{\textsc{Algorithmic Scenario Generation  \\ as Quality Diversity Optimization}
 \\\vspace{0.3cm}
 \large{Stefanos Nikolaidis} \\ \vspace{0.3cm}
\small Interactive and Collaborative Autonomous Robotic Systems (ICAROS) Lab \\ 
Thomas Lord Department of Computer Science\\
Viterbi School of Engineering \\ University of Southern California
\thanks{This is a summary of work done in collaboration with (in alphabetical order): Varun Bhatt, Bistra Dilkina, Ruilin Liu, Matthew Fontaine, Amy Hoover, Ya-chuan (Sophie) Hsu, Aniruddha Kalkar, Ahmed Khalifa, David Lee, Heramb Nemlekar, Anisha Palaparthi, Bryon Tjanaka, Julian Togelius, Hejia Zhang, Yulun Zhang. This work was supported by NSF CAREER \#2145077.}\\\vspace{0.1cm}\texttt{\small nikolaid@usc.edu}
    }

\begin{document}

\maketitle

\begin{abstract}
The increasing complexity of robots and autonomous agents that interact with people highlights the critical need for approaches that systematically test them before deployment. This review paper presents a general framework for solving this problem, describes the insights that we have gained from working on each component of the framework, and shows how integrating these components leads to the discovery of a diverse range of realistic and challenging scenarios that reveal previously unknown failures in deployed robotic systems interacting with people.
\end{abstract}

\section{Introduction}
\begin{figure}[t!]
\centering
\includegraphics[width=0.6\columnwidth]{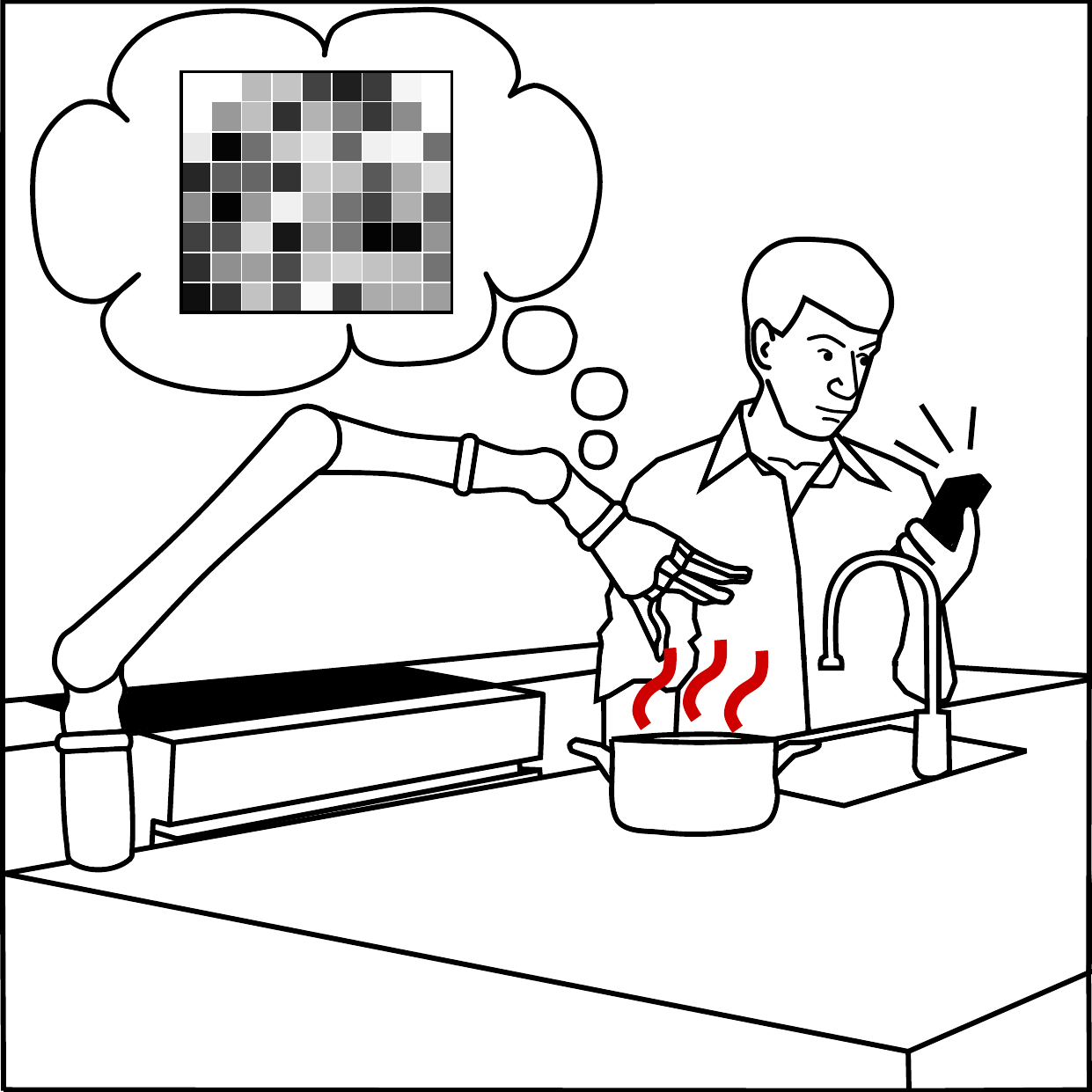}
\caption{The user gets distracted and forgets to empty the pot. Picking up the pot and attempting to place it on the stove would result in the robot dropping the pot with the boiling water. Our work on scenario generation can help avoid potentially catastrophic failures.}
\label{fig:hero}
\end{figure}

Consider a robot arm that collaborates with people in meal preparation tasks, such as steaming vegetables. After steaming, the user empties the boiling water into the nearby sink. The robot then places the pot back on the stove. We test the system with multiple participants in a user study and confirm that the human-robot team performed the task consistently well. We then deploy the robot at home, where the stove is far away from the sink. At one point during long-term deployment, the user gets distracted and forgets to empty the water (Fig.~\ref{fig:hero}). Unaware of that, the robot picks up the pot with the boiling water, but as it extends to reach the stove, it reaches its torque limits and drops the pot.

The complexity of interactions like this highlights the need for methods to systematically test novel algorithms and applications. Testing autonomous agents and robotic systems with actual users has limitations in the number of environments and user behaviors that can be covered. Exhaustive testing of every possible scenario in simulation is also computationally prohibitive. This highlights a critical need for generating scenarios in simulation that reveal undesirable behaviors.

This is a very challenging problem. It requires us to iteratively search the vast, continuous space of scenarios, generate realistic scenarios that are hard to find but may actually occur in the real world, and evaluate them by simulating agent and human behaviors. %

We address this research question by formulating the algorithmic scenario generation problem as a quality diversity optimization problem and proposing a general framework to solve it. This review paper highlights the insights that we have gained from our work on each component of the framework, building on previous advances on evolution strategies, generative modeling, surrogate models, and randomized algorithms. We demonstrate how integrating these components leads to the discovery of diverse, realistic and challenging edge-case failures in human-robot interaction. Finally, we  describe open scientific challenges that pave the way for future research.

\section{Problem Formulation} \label{sec:problem}

\noindent\textbf{Quality Diversity Optimization.} Quality diversity (QD) algorithms~\cite{pugh2015confronting} differ from pure optimization methods, in that they do not attempt to find a single optimal solution, but a collection of good solutions that differ across specified dimensions of interest, using existing solutions as ``stepping stones'' to generate higher quality and more diverse solutions.

Formally, the QD problem consists of an objective function  $f: \mathbb{R}^n \rightarrow \mathbb{R}^+$ that maps \mbox{$n$-dimensional} solution parameters to the scalar value representing the quality of the solution and $k$ measure functions, also commonly referred to as behavior descriptors, $m_i: \mathbb{R}^n \rightarrow \mathbb{R}$ or, as a vector function, $\bm{m}: \mathbb{R}^n \rightarrow \mathbb{R}^k$ that quantify the behavior or attributes of each solution. The range of $\bm{m}$ forms a measure space $S = \bm{m}(\mathbb{R}^n)$. The QD objective is to find a set of solutions $\bm{\theta} \in \mathbb{R}^n$, such that  $\bm{m}(\bm{\theta}) = \bm{s}$ for each $\bm{s}$ in $S$ and $f(\bm{\theta})$ is maximized.

The measure space $S$ is continuous, but solving algorithms need to produce a finite collection of solutions. Therefore, QD algorithms in the MAP-Elites~\cite{mouret2015illuminating,cully:nature15} family relax the QD objective by tessellating the space $S$ into $M$ cells that form an \textit{archive} $A$. Each solution $\vtheta_i$  is mapped to a cell in the archive based on its measure values $\bm{m}(\vtheta_i)$. The QD objective becomes to find a solution $\vtheta_i$ for each of the $i\in \{1,\ldots,M\}$ cells and to maximize the objective value $f(\vtheta_i)$ of all cells:

\begin{equation}
  \max\ J = \sum_{i=1}^M f(\bm{\theta_i})
\label{eq:qd_objective}
\end{equation}

\begin{figure}[t!]
\includegraphics[width=1.0\columnwidth]{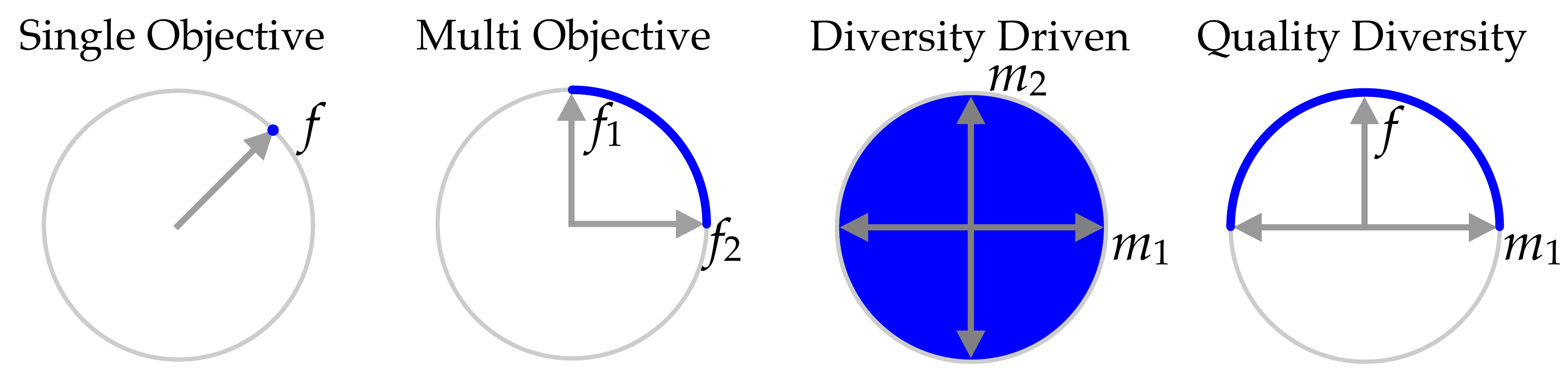}
\caption{QD and other types of optimization algorithms. The blue shading shows the desired result for each type.}
\label{fig:qd_shape}
\end{figure}

Fig.~\ref{fig:qd_shape} illustrates the differences between QD and other optimization approaches. Multi-objective optimization algorithms search for extreme points across two or more competing objectives $f_i$. Diversity-driven algorithms treat functions as measures and attempt to find a solution for each combination of outputs from the measure functions $\bm{m}$. QD algorithms optimize a single objective $f$ and can have multiple measures $\bm{m}$.

\noindent\textbf{Algorithmic Scenario Generation.} %
In algorithmic scenario generation, we wish to test and analyze the performance of one or more autonomous agents performing a task by executing a fixed policy.  The agents interact with the environment and possibly with other agents or humans. These interactions are governed by a set of scenario parameters $\vtheta$ that characterize the environment and the policies of the other agents or humans. 

We simulate the actions of all agents until task completion or a time limit. The evaluation returns measures $\bm{m}({\bm{\theta}})$ and a quality metric $f(\bm{\theta})$. Our objective is to find solutions $\bm{\theta}$ that fill in as many cells of the archive as possible with policies of high quality $f$. The measures $\bm{m}$ represent aspects of scenarios for which we wish to have diversity, such as scene clutter and human rationality. The objective $f$ represents the quality of the generated scenario. For instance, it can be the negative of a task performance metric when the goal is to identify failure scenarios, or the task performance metric itself when the aim is to find scenarios where the agents perform optimally.

Using the scenario of Fig.~\ref{fig:hero} as example, the robot executes a fixed policy and the scenario parameters $\vtheta$ include the object positions in the environment and the policy parameters of the simulated human. We assume as measure $m_1$ the user's observed rationality level, as $m_2$ the distance between the pot and the stove in the scene, and as objective $f$ the negative of task performance. Evaluating the scenario results in low $m_1$ and high $m_2$ values. It also results in a high $f$ value, because the team failed to successfully complete the task. The scenario will occupy a cell in the archive $A$ with coordinates ($m_1$, $m_2$).

\section{General Framework}
\begin{figure}[t!]
\centering
\includegraphics[width=0.8\columnwidth]{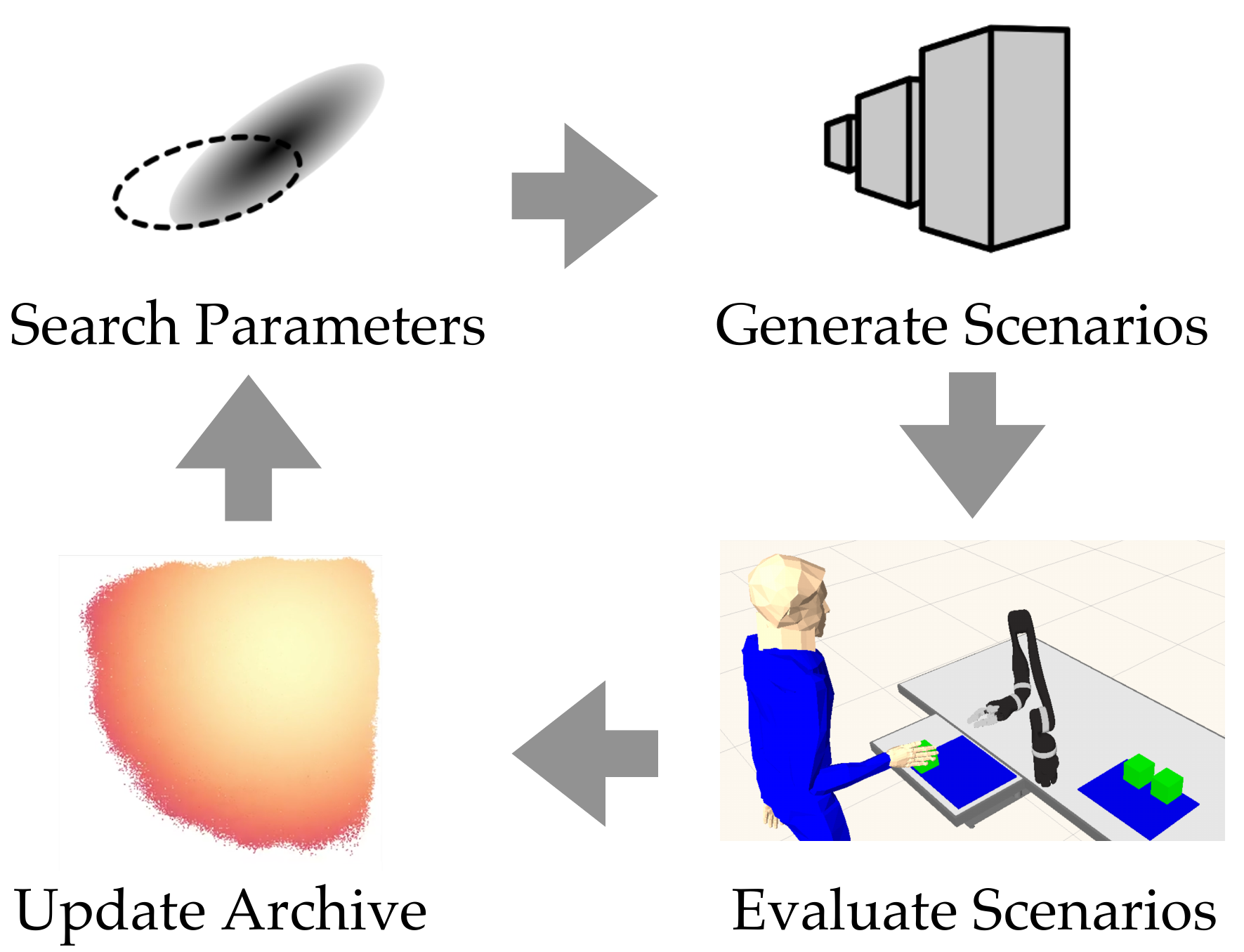}
\caption{Algorithmic scenario generation framework.}
\label{fig:framework}
\end{figure}

Our objective, as described in section~\ref{sec:problem}, is to populate an archive with high-quality scenarios that are diverse with respect to the specified measures of interest. Our framework consists of the following steps (Fig.~\ref{fig:framework}):
\begin{itemize}
\item Sample scenario parameters.
\item Generate scenarios in a simulator %
\item Evaluate the generated scenarios.
\item Update the archive with the new scenarios.%
\end{itemize}

We iterate the steps above for a maximum number of iterations or until we achieve a desired archive quality and coverage.

A naive approach would be to uniformly sample scenario parameters. In our running example of Fig.~\ref{fig:hero}, these would include the positions and the types of all objects at the scene and the parameters of the human policy. The next step would involve generating the scenario in a robotic simulator, executing a human and robot policy rollout and adding the new scenario to the archive. Unfortunately, this approach presents multiple challenges. 

\noindent\textbf{Measure Space Coverage.} First, uniformly sampling the high dimensional space of scenario parameters is unlikely to result in coverage of the lower-dimensional measure space. To provide some intuition, assume as measure the mean distance between objects and as scenario parameters the object positions. The distance of uniformly sampled objects will be concentrated tightly near the expected value~\cite{random2018}, and scenes where the objects are maximally or minimally distant will have zero probability. This implies the need for a targeted search that achieves greater coverage of the measure space.

\noindent\textbf{Scenario Realism.} In our running example, the pot may be sampled at a position that is unreachable by the robot or on top of the stove. The sampling should obey implicit and explicit constraints that exist in real-world scenarios. 

\noindent\textbf{Cost of Scenario Evaluation.} Third, evaluating a scenario in a simulator with stochastic agent policies and environment requires executing a large number of policy rollouts, which can be computationally expensive. At the same time, in the beginning of the search we often do not need the exact values of the objective and measures of the sampled scenarios, since these scenarios are likely to be replaced by higher-quality ones later on in the search.

\noindent\textbf{Scenario Registration.} Finally, filling the archive with every single scenario that is sampled would substantially increase the memory requirements. We need to consider methods that filter and prioritize which scenarios are stored in the archive, ensuring that the archive drives the search towards higher-quality, more diverse scenarios.

In the following sections, we describe how we have addressed each of these challenges.

\section{Search Scenario Parameters} \label{sec:search}
We describe how to efficiently search the continuous, multi-dimensional space of scenario parameters using QD optimization. 

\noindent\textbf{MAP-Elites.} A popular QD algorithm, \mapelites{}~\cite{mouret2015illuminating,cully:nature15}, generates an archive $A$ of high-performing solutions (scenario parameters) by retaining the best performing solution in each cell of the measure space. \mapelites{} first initializes the archive by sampling solutions from a Gaussian distribution. It then uniformly selects solutions $\bm{\theta}_i$ from the archive of solutions and perturbs them with isotropic Gaussian noise: 
\begin{equation}
\bm{\theta}'=  \bm{\theta}_i + \sigma \mathcal{N}(0,I) 
    \label{eq:mutation}
\end{equation}

\begin{figure}[t!]
\centering
\includegraphics[width=0.7\columnwidth]{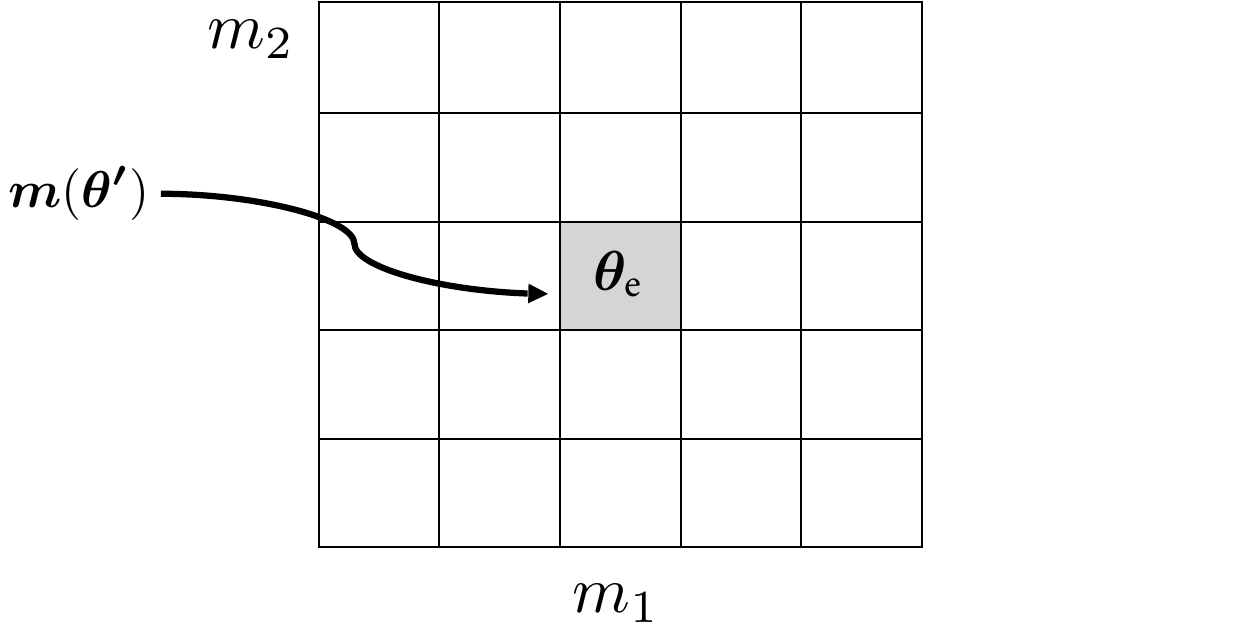}
\caption{The solution $\vtheta'$ is mapped to a cell $e$ in the archive based on the measure values $\bm{m}(\vtheta')$. The cell may be empty or occupied by an incumbent solution $\vtheta_e$}.
\label{fig:mapping}
\end{figure}

\mapelites{} evaluates each new solution and returns an objective value $f(\bm{\theta}')$ and measures $\bm{m}(\bm{\theta}')$. The algorithm then interacts with the archive by first mapping the solution $\vtheta'$ to a cell $e$ in the archive $A$ (Fig.~\ref{fig:mapping}). It then compares the objective value $f(\vtheta')$ of the new solution to a function $f_A(\vtheta')$ equal to the objective value of the incumbent solution in that cell, i.e., $f_A(\vtheta') = f(\vtheta_e)$. If  $\vtheta'$ has higher quality than $\vtheta_e$, i.e., $f(\vtheta') > f_A(\vtheta')$,  MAP-Elites replaces $\vtheta_e$ with $\vtheta'$. If cell $e$ is empty, we assume $f_A(\vtheta') = 0$ and $\vtheta'$ is added to the archive. This way, MAP-Elites retains an \textit{elitist} archive with each cell containing the highest performing solution found so far for that cell. The intuition behind MAP-Elites' iterative process is that the elites act as stepping stones to generate new, high-performing solutions in other parts of the archive.

While the simplicity of MAP-Elites makes it an attractive algorithm, a fundamental assumption when randomly mutating existing solutions (Eq.~\ref{eq:mutation}) is that every search direction is equally promising. However, this is not true in the general case. For example, if a region of the archive is saturated with high-quality solutions, it is beneficial to move the solution point in a direction that discovers new, previously unexplored cells in the archive. %

\noindent\textbf{CMA-ME.} Rather than applying random perturbations, we propose to move in the direction that maximizes the QD objective of Eq.~\ref{eq:qd_objective}. 

To specify this direction, we need to consider the change in the QD objective when a new solution $\bm{\theta'}$ is added to an  archive $A$. Before $\bm{\theta'}$  is added, the QD objective for archive $A$ is: $J(A)=\sum_{j=1}^M f(\bm{\theta}_j)$. After a new solution $\bm{\theta'}$ is added, the QD objective score improves as follows:

\begin{align}
\label{eq:improvement}
 J(A+\bm{\theta}') &=  \sum_{j=1}^N f(\bm{\theta}_j) + \Delta \\
 \text{where} \ \Delta &= f(\bm{\theta}') -  f_A(\vtheta') \notag
\end{align}

Our intuition is to \textit{explicitly search} for $\bm{\theta}'$ that maximizes archive improvement $\Delta = f(\bm{\theta}') -  f_A(\vtheta')$. We selected to use for the search Covariance Matrix Adaptation Evolution Strategies (CMA-ES)~\cite{hansen:cma16}, one of the best performing derivative-free optimizers. CMA-ES samples solutions from a multivariate Gaussian, selects and ranks them based on an objective, and updates the multivariate Gaussian based on the ranking. The self-adaptation and cumulation properties of CMA-ES allow it to robustly guide the algorithm toward promising regions of the search space. Previous work~\cite{akimoto2010bidirectional} has shown that CMA-ES approximates a natural gradient of the optimized objective.

We use CMA-ES to sample solutions $\bm{\theta}'$ from a multivariate Gaussian, but we select and rank them based on \textit{archive improvement} (Eq.~\ref{eq:improvement}), rather than based on objective $f$. %
This allows us to follow a natural gradient of the  QD objective. 

The resulting algorithm, Covariance Matrix Adaptation MAP-Elites (CMA-ME)~\cite{fontaine_GECCO_cmame}, is distinct from previous QD approaches in that it formulates the QD problem as an optimization problem that maximizes the QD objective. It combines the selection and ranking properties of CMA-ES with the archive interactions of MAP-Elites, achieving superior performance in multiple benchmark domains. 

\noindent\textbf{CMA-MEGA.} So far we have assumed no access to gradient information of the objective and measures. We refer to the problem where we have access to the exact gradients of the objective and measures as the differentiable quality diversity (DQD) problem.

The objective and measure gradients allow us to directly navigate the lower-dimensional objective-measure space, rather than the high-dimensional search space. As in CMA-ME, we can use CMA-ES to generate candidate solutions, but instead of sampling solutions directly in search space, we can leverage the gradient information to sample  gradient coefficients $\bm{c}$ and use the sampled coefficients to generate solutions $\bm{\theta}'$ in the lower-dimensional objective-measure space:
\begin{equation}
\bm{\theta}' = \bm{\theta} + c_0 \nabla f(\bm{\theta}) + \sum_{i=1}^k c_i \nabla m_i(\bm{\theta})
\label{eq:cma-mega}
\end{equation}

As in CMA-ME, we select and rank $\bm{\theta}'$ based on archive improvement and update the CMA-ES parameters towards the natural gradient of the QD objective. 

The resulting algorithm, Covariance Matrix Adaptation via Gradient Arborescence (CMA-MEGA)~\cite{fontaine_NeurIPS_dqd}, achieved orders of magnitude better performance than CMA-ME and MAP-Elites in benchmark domains.

We note that the problem of scenario generation is not a DQD problem; objective and measure gradients are typically not available, because of the agents' interaction with the environment and the often non-differentiable nature of the objective and measures. However, we can use approximations of the gradient information, as described in section~\ref{sec:update}.

\noindent\textbf{Key Insight.} \textit{We combine the archiving properties of MAP-Elites with the selection and ranking properties of CMA-ES to search for solutions that maximize archive improvement, thus following a natural gradient of the QD objective.}

\section{Generate Scenarios}
 \label{sec:generate}
 
Generating scenarios based on the sampled parameters often results in scenarios that are invalid or unrealistic. In our running example of Fig.~\ref{fig:hero}, all sampled positions of the pot should be reachable by the robot. Furthermore, there should be no decorative objects placed on top of or under the pot. 

\begin{figure}[t!]
\centering
\includegraphics[width=1.0\columnwidth]{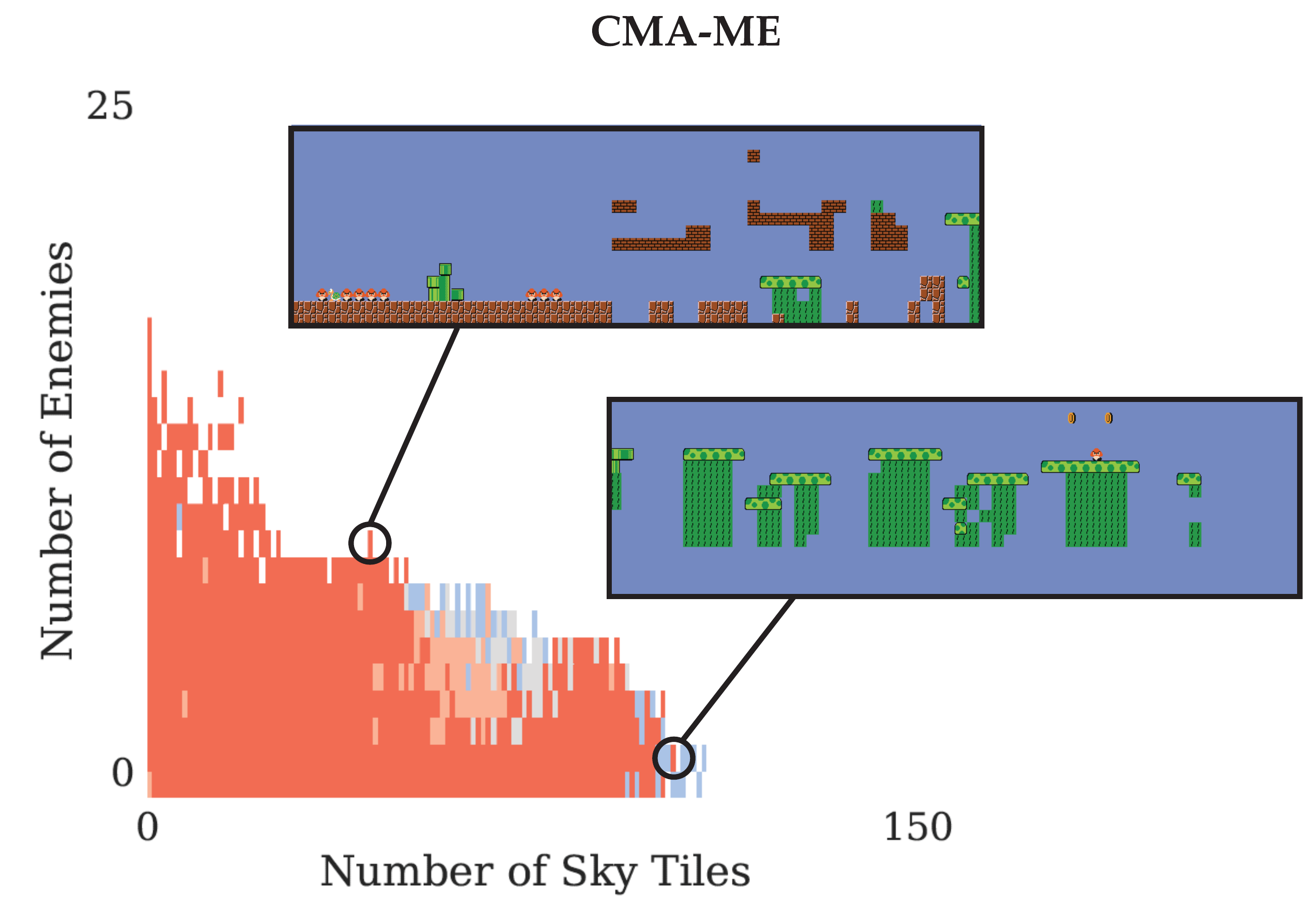}
\caption{Archive of generated Mario levels.}
\label{fig:mario}
\end{figure}

\noindent\textbf{Latent Space Illumination.} The use of generative models~\cite{goodfellow2020generative} has enabled the efficient generation of realistic content that matches the style of a training distribution. Furthermore, navigating the latent space of these models allows us to ``steer'' the output in a given direction. Our insight is that we can search with QD the latent space of generative models trained with human-authored examples to generate diverse scenarios that match the style of real-world examples. We call this approach \textit{latent space illumination}~\cite{fontaine_AAAI_lsi}. Fig.~\ref{fig:mario} shows an example application, where we search with CMA-ME the latent space of a GAN trained with human-authored Mario levels. The measures of diversity $\bm{m}$ are the number of sky tiles and number of enemies. The generated levels are diverse with respect to the measures of diversity, while resembling the human-authored levels.

One issue with latent space illumination is that environments produced by generative models are often invalid; in our running example of Fig.~\ref{fig:hero}, the robot's path to the pot may be blocked, or the scene may lack a pot altogether. In a Mario level, there may be no valid path to complete the level. 

To address this issue, we specified  as objective $f$ the percentage of the Mario level completed by an agent running an online A$^*$ algorithm, to guide the search towards valid levels. However, specifying a validity metric as objective presents the following challenges: (1) We may wish to specify other metrics of scenario quality, e.g., task performance, that are not aligned with the validity metric. (2) The search algorithm will often spend a large number of samples in generating invalid scenarios. Therefore, we propose letting the QD search generate environments freely and enforcing any validity constraints to the output of the generative model a posteriori.

\begin{figure}[t!]
\centering
\includegraphics[width=1.0\columnwidth]{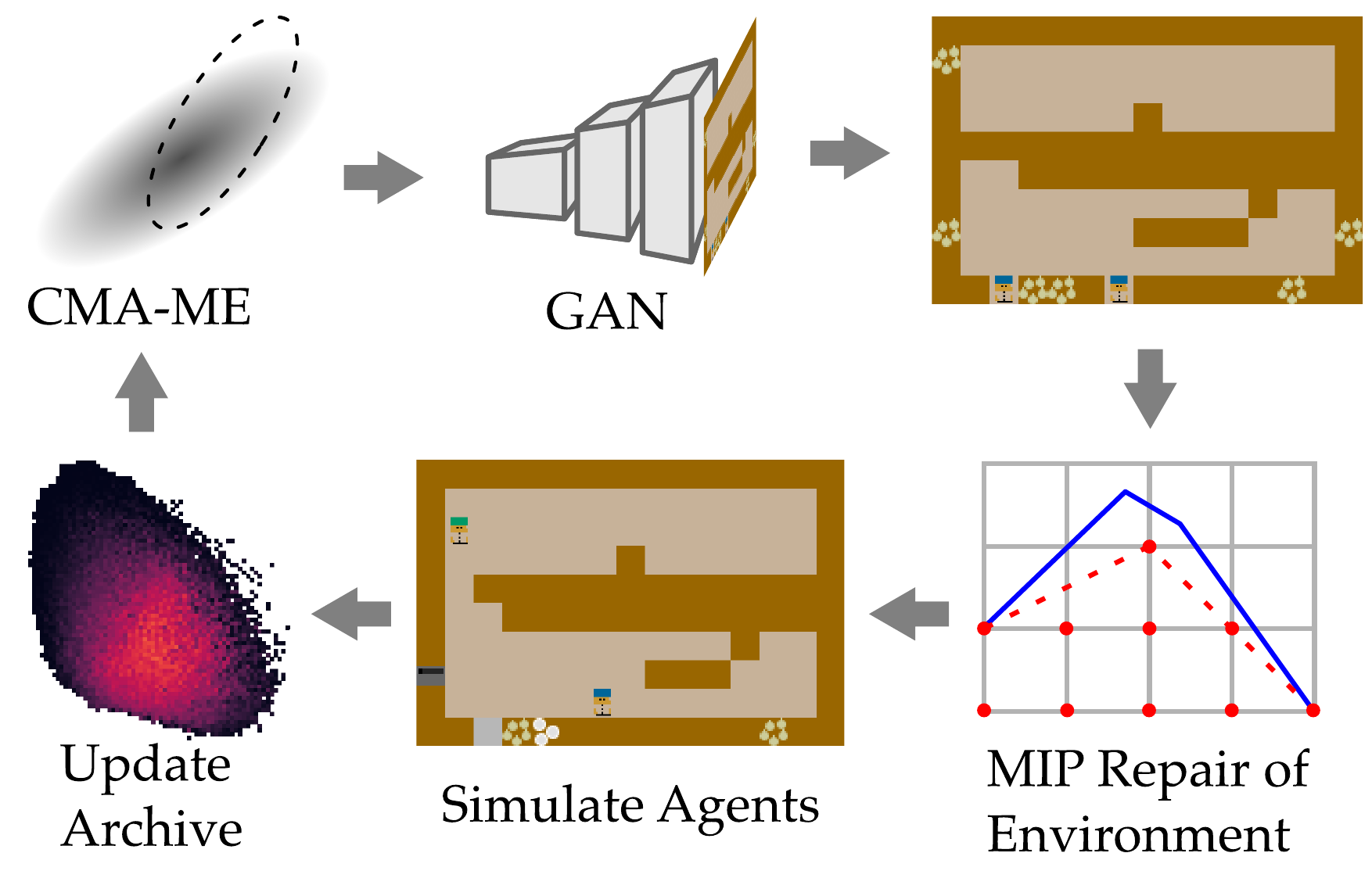}
\caption{We repair the invalid GAN-generated layouts with a MIP.}
\label{fig:overcooked-framework}
\end{figure}

\noindent\textbf{MIP Repair.}
 We propose editing the GAN-generated environments with a mixed-integer program (MIP), where the objective is to minimize the edit distance  to the GAN-generated environments~\cite{zhang_AIIDE_repair}, while satisfying prespecified constraints. Fig.~\ref{fig:overcooked-framework} shows an application of the proposed generate-and-repair method in the Overcooked game domain, where two agents collaborate to prepare a meal~\cite{fontaine2021importance}.

\begin{figure}[t!]
\includegraphics[width=0.49\columnwidth]{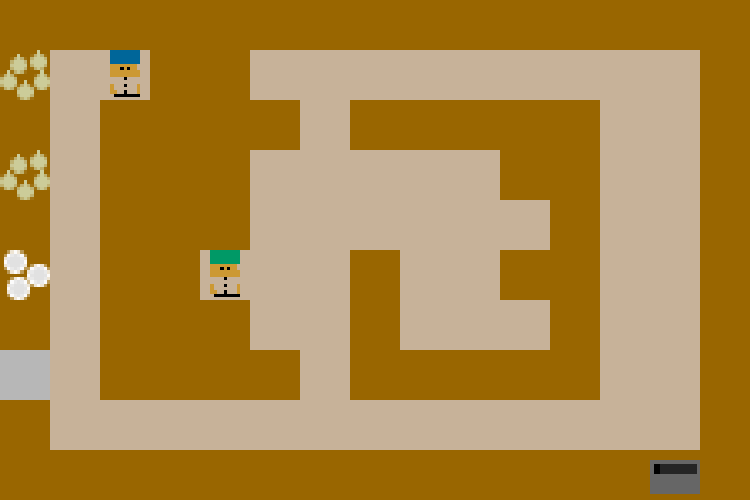}
\includegraphics[width=0.49\columnwidth]{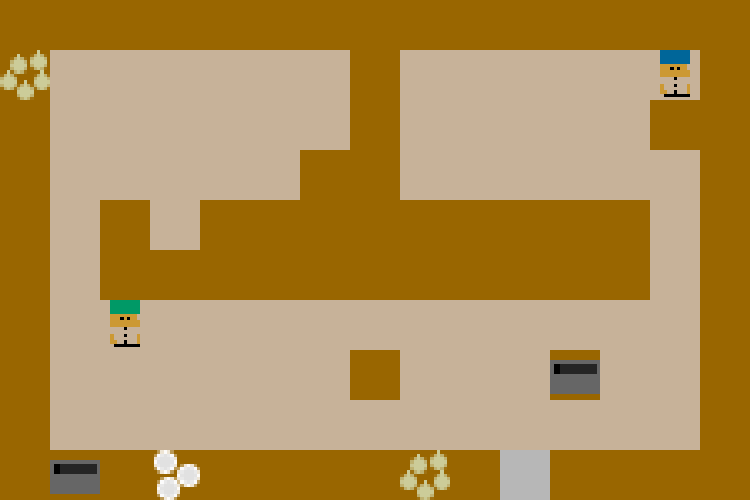}
\caption{Algorithmically generated \textit{Overcooked} layouts with low (left) and high (right) team fluency.}
\label{fig:overcooked}
\end{figure}

\begin{figure*}[t!]
\centering
\includegraphics[width=0.8\linewidth]{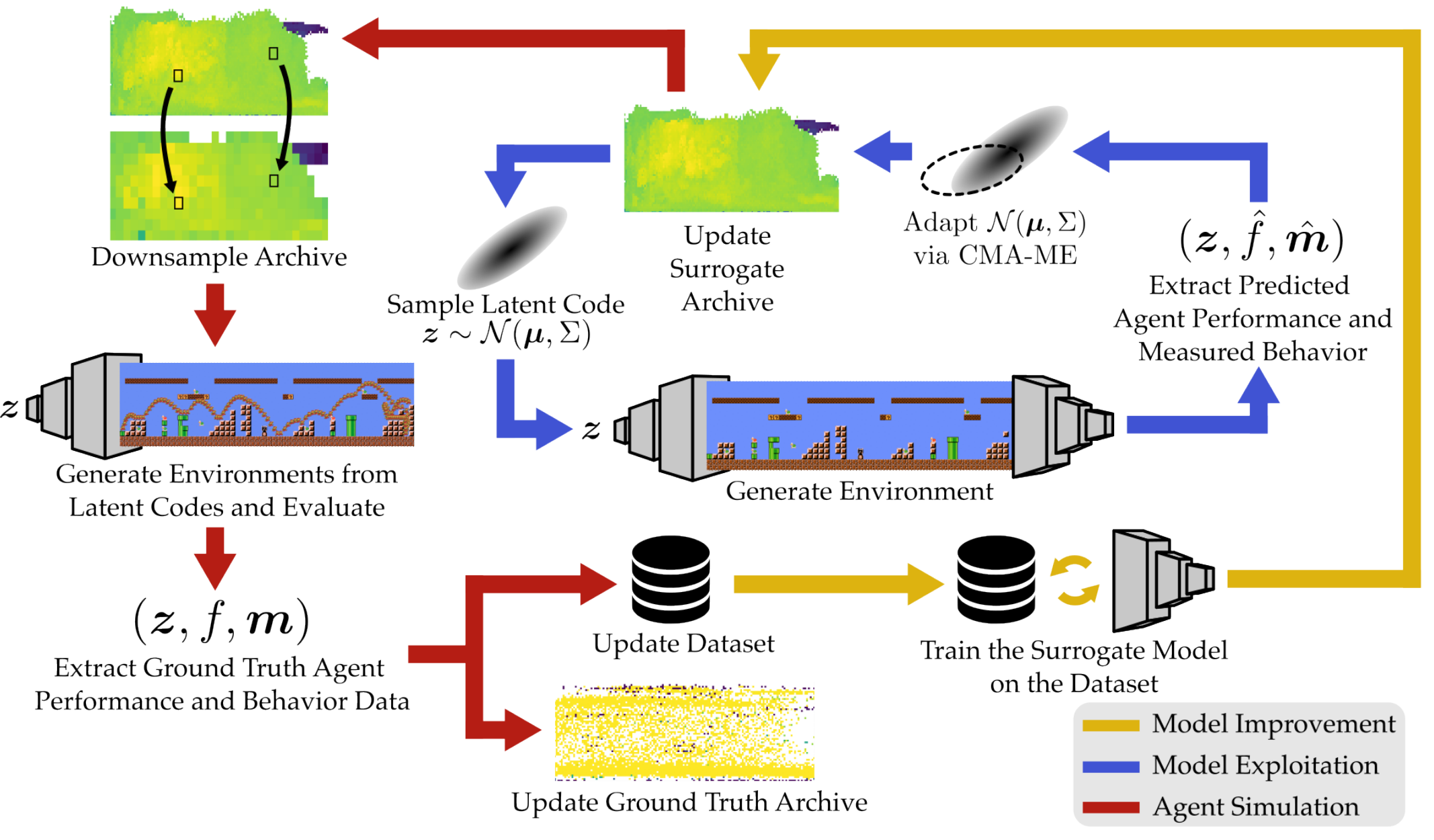}
 \caption{An overview of surrogate-based scenario evaluation. 
  An archive of solutions is generated by exploiting a deep surrogate model (\textbf{\color{Blue3}blue arrows}) with a QD optimizer, e.g., \cmame{}.
  We then downsample a subset of solutions from this archive and evaluate them by generating the corresponding environment and simulating an agent (\textbf{\color{Firebrick3}red arrows}). 
  The surrogate model is then trained on the data from the simulations (\textbf{\color{Goldenrod3}yellow arrows}).}
\label{fig:surrogate}
\end{figure*}

If we specify as measures of diversity metrics of team fluency~\cite{hoffman2019evaluating}, such as the percentage of concurrent motion and of the time the two agents get stuck, we can generate Overcooked layouts with high team fluency, as well as layouts where the agents struggle to coordinate. Fig.~\ref{fig:overcooked} shows two example generated layouts with low and high team fluency metrics. Fig.~\ref{fig:overcooked} (left) has all the objects of interest next to the blue agent on a narrow corridor, blocking access to the green agent. On the other hand, all the objects in Fig.~\ref{fig:overcooked} (right) are easily accessible, resulting in good coordination. While the generated layouts have human-like features, such as a long corridor in Fig.~\ref{fig:overcooked} (left), we did not explicitly search for layouts with these features. Instead, the search leverages the features learned by the generative model from the human-authored examples to generate human-like layouts that result in diverse team fluency metrics. 

\noindent\textbf{Key Insight.} \emph{By searching with QD the latent space of generative models trained with human-authored examples and repairing the output of the models with a MIP, we generate diverse, valid environments that retain the stylistic similarity of the human examples.}

\section{Evaluate Scenarios}
 \label{sec:evaluate}
Once the scenarios are generated, we evaluate them by executing the agent policies and computing the objective $f$ and measures $\textbf{m}$. Given the stochasticity of the policies and of the environment in the general case, evaluation requires multiple trials, which in a game engine or a robotics simulator can be time-consuming.

However, in the beginning of the QD search we  often do not need exact computations of the objective and measures, since the first solutions typically act as stepping stones, to be replaced by better solutions later on. If we could instead approximate the objective and measure values in the beginning of the search, we would avoid the expensive ground-truth evaluations.

One idea would be to train deep neural networks in a supervised learning manner so that they act as surrogate models, predicting the objective and measures given the scenario parameters. The predictions can be done efficiently through a forward pass. However, for the predictions to be accurate, we would need diverse training data. On the other hand, QD algorithms are specifically designed for the purpose of generating diverse data.  

We propose bringing the best of both worlds and co-train the deep neural networks with the QD search~\cite{zhang_GECCO_dsame,bhatt_NeurIPS_surrogate}. The algorithm consists of an inner loop and an outer loop (Fig.~\ref{fig:surrogate}). In the inner loop, we generate scenario parameters with the QD algorithm and use the networks as surrogate models. This results in the QD algorithm filling a \textit{surrogate archive}. In the beginning of the search, the untrained network fills the surrogate archive by making inaccurate predictions. However, the surrogate archive cells contain solutions that are diverse with respect to the predictions of the network. In the outer loop, we sub-sample the surrogate archive, label the sampled solutions with their ground-truth values by evaluating them in the simulator, add the labeled data to the training dataset, retrain the deep neural network, and repeat the process. We simultaneously leverage the ground-truth evaluations to fill in a ground-truth archive, which is the final output of the algorithm. Over time, \textit{the surrogate model learns to correct its own errors and gradually becomes more and more accurate, guiding the QD search towards diverse, high-quality solutions}.

We applied this algorithm to predict the performance of planning-based agents in the Mario and Overcooked game domains, and of reinforcement learning agents in a Maze domain. We have observed that using the surrogate model results in substantial improvements in the quality and diversity of the solutions. In addition to the objective and measures, having the network predict \textit{ancillary agent behavior data}, in the form of an occupancy grid that specifies the number of times an agent will visit each tile in the environment, was essential in achieving good performance.

Analysis of the results has shown that an important part of our implementation is resetting the surrogate archive after each inner loop evaluation. The reason is that as the network changes after each outer-loop iteration, each corresponding surrogate archive tends to occupy different cells of the ground-truth archive, and the union of these archives results in dense coverage. This result is consistent with previous insights on the benefits of the mortality on the evolvability of the population in divergent search algorithms~\cite{lehman2015enhancing}.

\noindent\textbf{Key Insight.} \emph{By exploiting a surrogate model with QD, we obtain diverse scenarios to train a better surrogate model, and a better  model helps guide the QD search towards high-quality and diverse scenarios.}

\begin{figure}[t!]
\centering
\includegraphics[width=1.0\columnwidth]{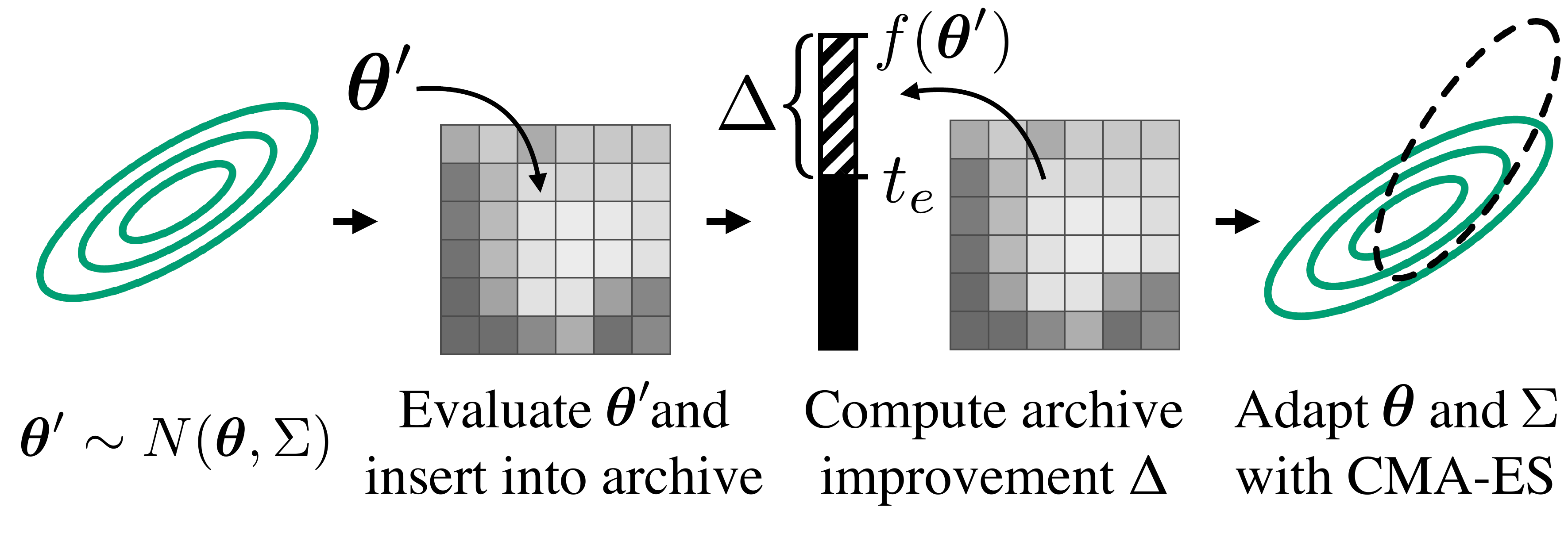}
\caption{The CMA-MAE algorithm.}
\label{fig:cma_mae}
\end{figure}

\begin{figure*}[!t]
\centering
\includegraphics[width=0.8\linewidth]{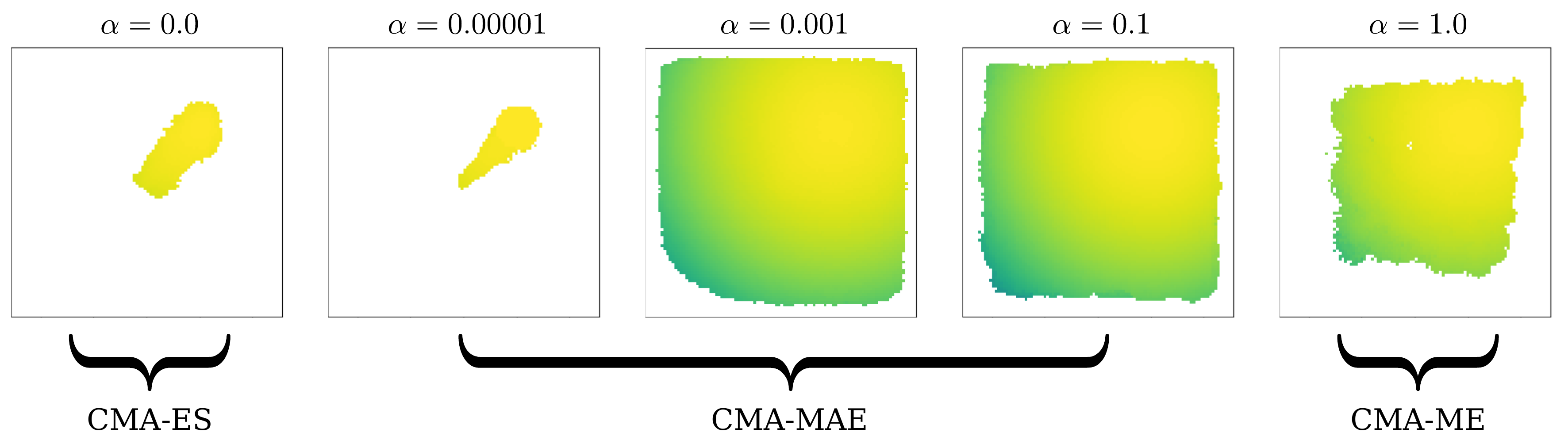}
\caption{Our proposed CMA-MAE algorithm smoothly blends between the behavior of CMA-ES and CMA-ME via an archive learning rate $\alpha$. Each heatmap visualizes an archive of solutions across a 2D measure space, where the color of each cell represents the objective value of the solution.}
\label{fig:blend}
\end{figure*}

\section{Update Archive}
 \label{sec:update}
The final component of the framework is to add the new scenarios to the archive. A key observation is that how solutions are added to the archive has a significant effect on the performance of the search, since QD algorithms interact with the evolving archive throughout the search. 

We recall that CMA-ME (section~\ref{sec:search})  determines a search direction by sampling solutions from a multi-variate Gaussian and ranking the sampled solutions based on archive improvement. The archive improvement of a sampled solution $\vtheta'$ is computed as $\Delta = f(\vtheta')-f_A(\vtheta')$. As we describe in section~\ref{sec:search}, in elitist archives $f_A(\vtheta') = f(\vtheta_e)$ with $\vtheta_e$ the best performing solution found at cell $e$. CMA-ME then moves in the direction of higher-ranked solutions.

\noindent\textbf{Elitist Archive.} We provide an example of how CMA-ME can behave when interacting with an elitist archive. Assume that CMA-ME samples a new solution $\vtheta'$ with a high objective value of $f(\vtheta') = 99$. If the current occupant $\vtheta_e$ of the corresponding cell has a low objective value of $f(\vtheta_e) = 0.3$, then the improvement in the archive $\Delta = f(\vtheta') - f(\vtheta_e) = 98.7$ is high and the search will move towards $\vtheta'$. Now, assume that in the next iteration the algorithm discovers a new solution $\vtheta''$ with objective value $f(\vtheta'') = 100$ that maps to the same cell as $\vtheta'$. The improvement then is $\Delta = f(\vtheta'') - f(\vtheta') = 1$ as $\vtheta'$ replaced $\vtheta_e$ in the archive in the previous iteration. CMA-ME would likely move away from $\vtheta''$ because the solution resulted in low improvement. In contrast, the single-objective optimizer CMA-ES would move towards $\vtheta''$ because it ranks only by the objective $f$, ignoring previously discovered solutions with similar measure values.

In the above example, CMA-ME moves away from a high performing region in order to maximize how the archive changes. However, in domains with hard-to-optimize objective functions, it is beneficial to perform more optimization steps towards the objective $f$ before leaving each high-performing region.

\noindent\textbf{Soft Archive.} We present a new algorithm, CMA-MAE~\cite{fontaine2022covariance}, that addresses this limitation. Like CMA-ME, CMA-MAE maintains a discount function $f_A(\vtheta')$ and ranks solutions by improvement $f(\vtheta')-f_A(\vtheta')$. However, instead of maintaining an \textit{elitist archive} by setting $f_A(\vtheta')$ equal to $f(\vtheta_e)$, we maintain a \textit{soft archive} by setting $f_A(\vtheta')$ equal to $t_e$, where $t_e$ is an acceptance threshold maintained for each cell in the archive (Fig.~\ref{fig:cma_mae}).  When adding a candidate solution $\vtheta'$ to the archive, we control the rate that $t_e$ changes by the archive learning rate $\alpha$ as follows: $t_e \leftarrow (1-\alpha) t_e + \alpha f(\vtheta')$.

The archive learning rate $\alpha$ in CMA-MAE allows us to control how quickly we leave a high-performing region of measure space. For example, consider discovering solutions in the same cell with objective value 100 in 5 consecutive iterations. The improvement values computed by CMA-ME against the elitist archive would be $100,0,0,0,0$, thus CMA-ME would move rapidly away from this cell. The improvement values computed against the soft archive of CMA-MAE with $\alpha=0.5$ would diminish smoothly as follows:  $100,50,25,12.5,6.25$, enabling further exploitation of the high-performing region. Furthermore, if $\alpha=0$, we ignore the archive and CMA-MAE  is equivalent to the single-objective optimizer CMA-ES. If $\alpha=1$, we set the threshold to the objective of the new incumbent solution, and CMA-MAE is equivalent to CMA-ME. For $0 < \alpha < 1$, we smoothly blend between behavior equivalent to CMA-ES and behavior equivalent to CMA-ME (Fig.~\ref{fig:blend}). 

The annealing of the acceptance threshold comes with an additional benefit. When running CMA-ME on  flat objective regions, where all solutions have the same objective $f(\vtheta)=C$, the archive will change once when populating the empty cells with solutions and then will stop changing. Thus, CMA-ME will not have any signal on which direction to explore. On the other hand, in flat objectives the acceptance threshold in CMA-MAE will increase proportionally to the number of times each cell is visited. Thus, cells that are visited less frequently will have smaller threshold values. This leads CMA-MAE to behave as a density descent on the visitation frequency histogram of the archive, moving towards the less frequently visited cells and escaping the flat objective region.

These benefits of CMA-MAE have significant implications in scenario generation. The ability to optimize high-performing regions enables the search to discover rare, hard-to-find edge cases. Furthermore, failure scenarios, e.g., scenarios where the team fails to complete the task within a maximum time limit, have the same objective value $f$. It is critical for the algorithm to escape these flat objective regions to discover diverse types of failures. 

Our augmentations to the CMA-ME algorithm only affects how we replace solutions in the archive and how we calculate the archive improvement $\Delta$. CMA-ME and CMA-MEGA replace solutions and calculate $\Delta$ identically, thus we have applied the same augmentations to CMA-MEGA to form a new DQD algorithm, CMA-MAEGA~\cite{fontaine2022covariance}. 

\noindent\textbf{Key Insight.} \textit{By introducing an acceptance threshold in each cell of the archive and annealing it with a learning rate, we can smoothly blend between the behavior of CMA-ES and CMA-ME.}

\section{Case Study}
The previous sections have discussed how to efficiently search the continuous space of scenario parameters (section~\ref{sec:search}), how to generate realistic and valid scenarios (section~\ref{sec:generate}), how to efficiently evaluate the generated scenarios (section~\ref{sec:evaluate}) and how to update the archive of scenarios to enable the search algorithm to focus on high-performing solutions (section~\ref{sec:update}). Here we discuss how we can leverage our contributions in all the components of the framework to find diverse failure scenarios in human-robot interaction. 

\begin{figure}[!t]
\centering
\includegraphics[width=0.8\linewidth]{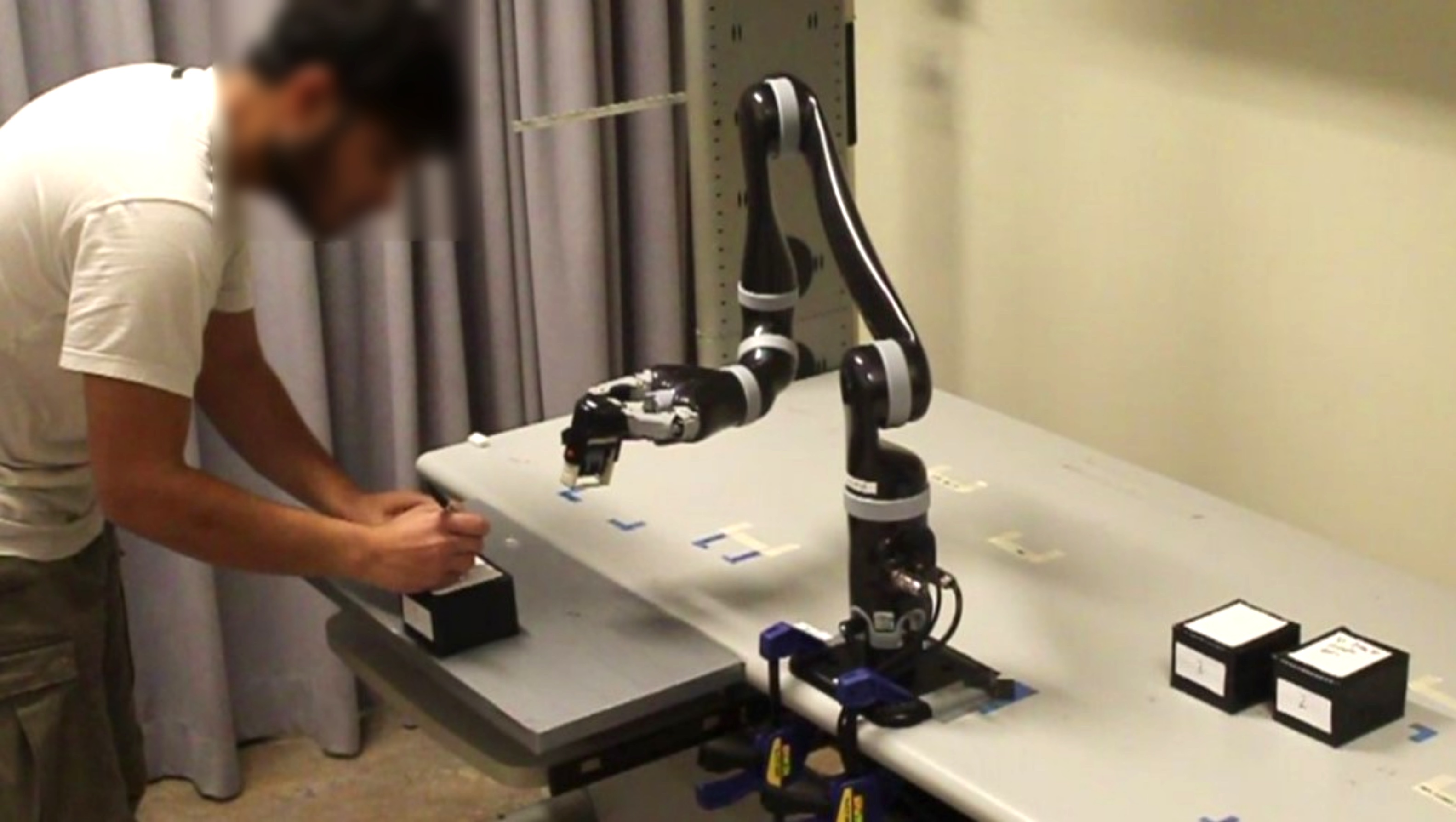}
\caption{The shared workspace collaboration task.}
\label{fig:collab_task_real}
\end{figure}

We consider two different domains: (1) A shared workspace collaboration task, where a user attaches a label on one of three packages while the robot presses a stamp (\fref{fig:collab_task_real}).  The robot infers the human's current goal and moves to a different goal along the shortest path.  (2) A shared control teleoperation task, where the user provides inputs to the robot through a joystick interface. The robot uses the inputs to infer the human goal and moves along the shortest path to that goal. In both domains, the team attempts to complete the task as quickly as possible.

We parameterize the scenario with the coordinates of the goal objects and the parameters of the human model, such as the type of each human action and the speed of the human movements. Our objective is to find scenarios that minimize the performance of the team. We select as measures of diversity factors that affect the type of scenarios that we wish to explore; for instance, by selecting as measure the maximum probability assigned by the robot to a goal other than the true human goal, we search for failure scenarios that are caused by incorrect goal inference, but also for failure scenarios where the robot inferred the human goal correctly. By selecting as measure the human rationality, we simulate human actions that are optimal, but also actions that are completely random. 

Following the proposed framework, we can use QD to search for scenario parameters that maximize the objectives and fill the archive specified by the measure space (section~\ref{sec:search}). We use CMA-MAE and anneal the acceptance thresholds with a low value of the learning rate $\alpha = 0.1$ to guide the search towards the high-performing solutions (section~\ref{sec:update}). Rather than evaluating all scenarios on the ground-truth simulator, we use a deep neural network as a surrogate model that approximates the human-robot interaction outcomes, generating a surrogate archive (section~\ref{sec:evaluate}). Furthermore, since the surrogate model can backpropagate gradient information of the objective and measures, we use CMA-MAEGA, the differentiable quality diversity variant of CMA-MAE.

\begin{figure}[!t]
\centering
\includegraphics[width=1.0\linewidth]{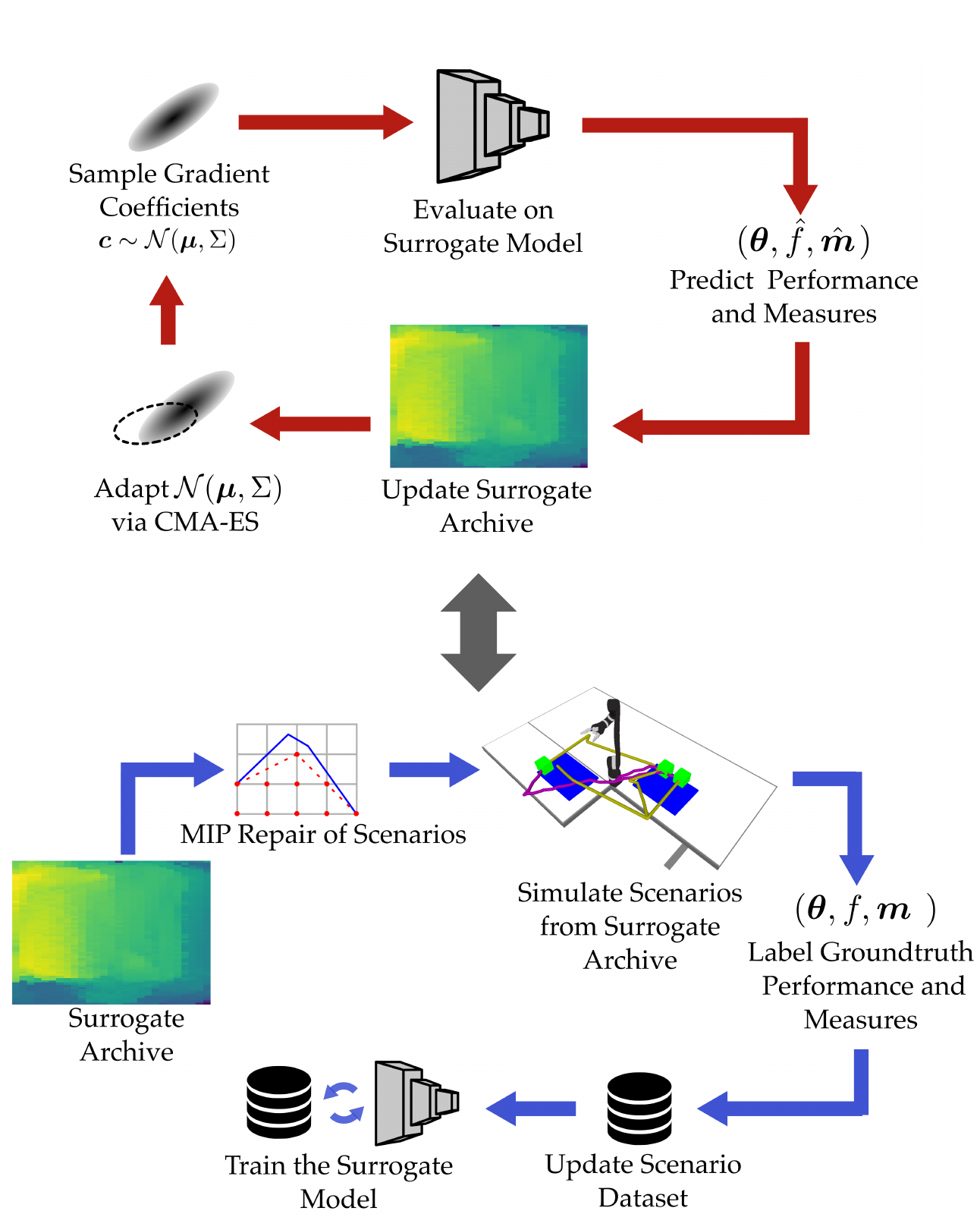}
\caption{Generating HRI scenarios with the proposed framework. We alternate between exploiting a surrogate model that predicts the outcome of the human and robot interaction (\textbf{\color{Firebrick3}red arrows}), and evaluating candidate scenarios on the simulator and adding them to the dataset (\textbf{\color{Blue3}blue arrows}).}
\label{fig:hri_framework}
\end{figure}

We then subsample the surrogate archive, repair the generated scenarios with a Mixed Integer Program (section~\ref{sec:generate}) to ensure validity and evaluate them in the ground-truth simulator. We use the labeled data to retrain the deep neural network and repeat the process (Fig.~\ref{fig:hri_framework}).

Our scenario search results in rare, hard-to-find edge cases that would be impossible to anticipate without a systematic approach~\cite{bhatt_CORL_surrogate}. For instance, in the workspace collaboration task, we find object arrangements that result in significant delays in task completion time. In one of the scenarios, the simulated human goes to one of the goal objects too fast for the robot to infer their goal accurately, which makes the robot move to the same goal object as the human. In another scenario, the robot infers the human goal correctly but the object arrangement results in a robot motion towards its joint-limits, forcing the robot to re-plan. There are also scenarios where the human and the robot operate on different objects, but they end up needing to work on the same object to complete the task, resulting in a long wait time.\footnote{Videos of the failure scenarios are in \url{https://youtu.be/AU2MUxbTeCo}}

\begin{figure}[t!]
\centering
\includegraphics[width=0.8\columnwidth]{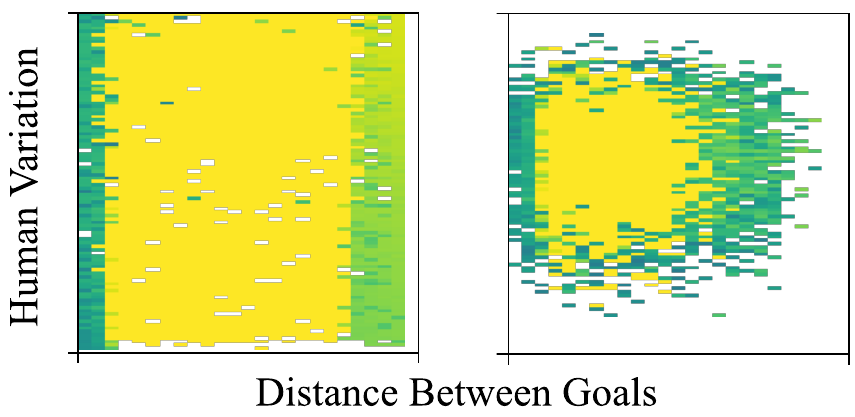}
\caption{Archive of failure scenarios in the shared control teleoperation domain using the proposed framework (left) and uniform sampling (right).}
\label{fig:heatmaps}
\end{figure}

An important benefit of our approach is that visualizing the 2D and 3D archives provides qualitative insights about the behavior of the algorithm. We use as example the archive generated in the shared control teleoperation domain with measures the human variation (total amount of noise between human inputs) and the mean distance between objects (Fig.~\ref{fig:heatmaps}). Yellow cells indicate timeout failures, where the robot was unable to reach the goal within a time-limit. 

We observe that if the distance between the goal objects is above a threshold, there are no time-outs. The reason is that the objects are sufficiently far apart for the robot to infer correctly the human goal, regardless of how noisy the human inputs are. On the other hand, even for low measure values of human variation, where the simulated human is nearly optimal, the QD search finds object arrangements that result in failure. Fig.~\ref{fig:heatmaps}
(right) shows that if instead of using the proposed approach, we uniformly sampled object locations and human actions within the same domain, we would achieve only limited archive coverage  which would restrict our ability to derive meaningful insights.

\section{Looking Ahead}
We have presented a general framework for algorithmic scenario generation. We have shown how combining the archiving properties of MAP-Elites with the ranking and selection properties of CMA-ES enables us to efficiently explore the vast, continuous space of scenarios. Integrating our QD algorithms with generative models trained with human-authored examples allows us to generate valid, realistic scenarios. Using deep neural networks as surrogate models for evaluation improves sample efficiency. Finally, introducing an acceptance threshold into each cell of the archive guides the search towards high-performing solutions and helps it escape flat objective regions. Our framework can find rare, previously unknown failures in human-robot interaction domains.

Our work has a number of limitations that suggest potential future directions.

\noindent\textbf{Scenario Complexity.} As we increase the complexity of the scenes and the number of agents acting in the environment, the number of scenario parameters will also increase. Our QD algorithms are based on CMA-ES, which has quadratic time complexity to the number of parameters.  CMA-ES’s complexity arises from its modeling of the search distribution with a Gaussian that has a full rank
covariance matrix. This naturally limits the maximum number of scenario parameters.

We have explored CMA-MAE variants that approximate the covariance matrix with a low-dimensional representation or with a diagonal matrix~\cite{tjanaka_RAL_scaling}. We have found that in many domains these variants achieve performance close to that of CMA-MAE, with orders of magnitude improvements in computational complexity. 

Alternatively, if the objective and measure gradients are available or can be approximated, DQD algorithms such as CMA-MAEGA retain a covariance matrix in the much lower-dimensional space of the objective-measure gradient coefficients (Eq.~\ref{eq:cma-mega}), instead of the space of scenario parameters, and do not suffer from scenario space scalability limitations. We are excited about future work that leverages these insights to improve the complexity of the generated scenarios.

\noindent\textbf{Archive Size.} We have discussed that tessellating the measure space results in an archive of solutions, which consists of uniformly spaced grid cells. The number of cells thus increases exponentially to the number of measure functions, leading to a ``curse of dimensionality.'' 

One approach would be to specify first the number of desired cells and then maximally spread them in the measure space, e.g., by using a centroidal Voronoi tessellation (CVT) of that space~\cite{vassiliades2017using}. However, this would significantly increase the volume of each cell with the number of dimensions. In turn, this can cause the solutions sampled by the QD algorithm to all land in the same cell, heavily impacting its ability to explore new cells. 

Rather than relying on measure space tessellations, we have proposed using continuous density estimation of the measure space~\cite{Lee_GECCO_Density}. In the special case where we focus only on diversity (diversity optimization) and ignore solution quality, we have observed notable performance in high-dimensional measure spaces. We look forward to future work that uses continuous density estimation to solve the quality diversity problem.

\noindent\textbf{Scene Realism.} While we have applied latent space illumination on GANs to generate diverse, realistic scenes, diffusion models~\cite{sohl2015deep} have emerged as a current state-of-the-art approach in generative modeling, considering their capability to generate high-resolution, complex scenes that capture the natural variations found in real-world scenes. A recent tutorial~\cite{pyribs_diffusion} that was introduced in our open-source QD library \textit{pyribs}~\cite{tjanaka2023pyribs} describes the integration of QD algorithms with diffusion models. As more powerful generative models continue to emerge, we anticipate illuminating the latent space of these models to further enhance the realism of the generated scenarios.

\noindent\textbf{Human Model Realism.} We have shown how integrating QD algorithms with generative models of environments can result in realistic environments. We expect these insights to transfer to the simulation of realistic human behaviors. For instance, in our past work~\cite{nikolaidis2017mathematical}, we have modeled human decision making using learned low-dimensional representations of human strategies and Theory-of-Mind models. We believe that learning compact representations of human states and actions and searching with QD these representations holds much promise. 

\noindent\textbf{Robustness.} Ultimately, the purpose of generating failure scenarios is to enhance the robustness of deployed robotic systems. We have ourselves leveraged the generated failure scenarios in the shared control teleoperation example (Fig.~\ref{fig:heatmaps}) to refine the design of the robot's cost function and prevent it from failing~\cite{fontaine2021iquality}.

Beyond manual debugging, the generated archive of scenarios serves as a diverse, high-quality dataset that could provide a foundation for training learning algorithms. We hypothesize that leveraging the structure of the dataset can significantly improve the learning efficiency.

\noindent\textbf{Conclusion.} While we have presented single and multi-agent games, as well as human-robot collaborative tasks as application domains, our algorithmic scenario generation framework is general; for instance, recent works have algorithmically generated diverse prompts for red-teaming LLMs~\cite{samvelyan2024rainbow} and diverse safety-critical scenarios for autonomous cars~\cite{huang2024cadre}. We expect our findings to have significant impact in other domains as well.

Overall, the presented work reflects our long-term vision that deployed autonomous agents and robotic systems can continuously improve their robustness through simulated and real-world experiences. We look forward to continue addressing the
exciting scientific challenges in this area.

\newpage

\bibliographystyle{unsrt}
\bibliography{conferences}

\begin{thebibliography}{10}

\bibitem{pugh2015confronting}
Justin~K Pugh, Lisa~B Soros, Paul~A Szerlip, and Kenneth~O Stanley.
\newblock Confronting the challenge of quality diversity.
\newblock In {\em Proceedings of the 2015 Annual Conference on Genetic and
  Evolutionary Computation}, pages 967--974, 2015.

\bibitem{mouret2015illuminating}
Jean-Baptiste Mouret and Jeff Clune.
\newblock Illuminating search spaces by mapping elites.
\newblock {\em arXiv preprint arXiv:1504.04909}, 2015.

\bibitem{cully:nature15}
Antoine Cully, Jeff Clune, Danesh Tarapore, and Jean-Baptiste Mouret.
\newblock Robots that can adapt like animals.
\newblock {\em Nature}, 521(7553):503, 2015.

\bibitem{random2018}
Roman Vershynin.
\newblock {\em Random Vectors in High Dimensions, page 3869}.
\newblock Cambridge Series in Statistical and Probabilistic Mathematics.
  Cambridge University Press, 2018.

\bibitem{hansen:cma16}
Nikolaus Hansen.
\newblock The cma evolution strategy: A tutorial.
\newblock {\em arXiv preprint arXiv:1604.00772}, 2016.

\bibitem{akimoto2010bidirectional}
Youhei Akimoto, Yuichi Nagata, Isao Ono, and Shigenobu Kobayashi.
\newblock Bidirectional relation between cma evolution strategies and natural
  evolution strategies.
\newblock In {\em International Conference on Parallel Problem Solving from
  Nature}, pages 154--163. Springer, 2010.

\bibitem{fontaine_GECCO_cmame}
Matthew Fontaine, Julian Togelius, Stefanos Nikolaidis, and Amy Hoover.
\newblock Covariance matrix adaptation for the rapid illumination of behavior
  space.
\newblock In {\em Proceedings of the 2020 Genetic and Evolutionary Computation
  Conference (GECCO)}, 2020.

\bibitem{fontaine_NeurIPS_dqd}
Matthew Fontaine and Stefanos Nikolaidis.
\newblock Differentiable quality diversity.
\newblock In {\em Neural Information Processing Systems (NeurIPS)}, 2021.

\bibitem{goodfellow2020generative}
Ian Goodfellow, Jean Pouget-Abadie, Mehdi Mirza, Bing Xu, David Warde-Farley,
  Sherjil Ozair, Aaron Courville, and Yoshua Bengio.
\newblock Generative adversarial networks.
\newblock {\em Communications of the ACM}, 63(11):139--144, 2020.

\bibitem{fontaine_AAAI_lsi}
Matthew Fontaine, Ruilin Liu, Ahmed Khalifa, Jignesh Modi, Julian Togelius, Amy
  Hoover, and Stefanos Nikolaidis.
\newblock Illuminating mario scenes in the latent space of a generative
  adversarial network.
\newblock In {\em 35th AAAI Conference on Artificial Intelligence (AAAI)},
  2021.

\bibitem{zhang_AIIDE_repair}
Hejia Zhang, Matthew Fontaine, Amy Hoover, Julian Togelius, Bistra Dilkina, and
  Stefanos Nikolaidis.
\newblock Video game level repair via mized integer linear programming.
\newblock In {\em Proceedings of the AAAI Conference on Artificial Intelligence
  and Interactive Digital Entertainment (AIIDE)}, 2020.

\bibitem{fontaine2021importance}
Matthew~C Fontaine, Ya-Chuan Hsu, Yulun Zhang, Bryon Tjakana, and Stefanos
  Nikolaidis.
\newblock On the importance of environments in human-robot coordination.
\newblock In {\em Robotics Science and Systems (RSS)}, 2021.

\bibitem{hoffman2019evaluating}
Guy Hoffman.
\newblock Evaluating fluency in human--robot collaboration.
\newblock {\em IEEE Transactions on Human-Machine Systems}, 49(3):209--218,
  2019.

\bibitem{zhang_GECCO_dsame}
Yulun Zhang, Matthew Fontaine, Amy Hoover, and Stefanos Nikolaidis.
\newblock Dsa-me: Deep surrogate assisted map-elites.
\newblock In {\em The Genetic and Evolutionary Computation Conference (GECCO)},
  2022.

\bibitem{bhatt_NeurIPS_surrogate}
Varun Bhatt, Bryon Tjanaka, Matthew Fontaine, and Stefanos Nikolaidis.
\newblock Deep surrogate assisted generation of environments.
\newblock In {\em Neural Information Processing Systems (NeurIPS)}, 2022.

\bibitem{lehman2015enhancing}
Joel Lehman and Risto Miikkulainen.
\newblock Enhancing divergent search through extinction events.
\newblock In {\em Proceedings of the 2015 Annual Conference on Genetic and
  Evolutionary Computation}, pages 951--958, 2015.

\bibitem{fontaine2022covariance}
Matthew~C Fontaine and Stefanos Nikolaidis.
\newblock Covariance matrix adaptation map-annealing.
\newblock {\em The Genetic and Evolutionary Computation Conference (GECCO)},
  2023.

\bibitem{bhatt_CORL_surrogate}
Varun Bhatt, Heramb Nemlekar, Matthew Fontaine, Bryon Tjanaka, Hejia Zhang,
  Ya-Chuan Hsu, and Stefanos Nikolaidis.
\newblock Surrogate assisted generation of human-robot interaction scenarios.
\newblock In {\em Conference on Robot Learning (CORL)}, 2023.

\bibitem{tjanaka_RAL_scaling}
Bryon Tjanaka, Matthew Fontaine, Aniruddha Kalkar, and Stefanos Nikolaidis.
\newblock Training diverse high-dimensional controllers by scaling covariance
  matrix adaptation map-annealing.
\newblock In {\em Robotics and Automation Letters (RA-L)}, 2023.

\bibitem{vassiliades2017using}
Vassilis Vassiliades, Konstantinos Chatzilygeroudis, and Jean-Baptiste Mouret.
\newblock Using centroidal voronoi tessellations to scale up the
  multidimensional archive of phenotypic elites algorithm.
\newblock {\em IEEE Transactions on Evolutionary Computation}, 22(4):623--630,
  2017.

\bibitem{Lee_GECCO_Density}
David Lee, Anisha Palaparthi, Matthew Fontaine, and Stefanos Nikolaidis.
\newblock Density descent for diversity optimization.
\newblock In {\em The Genetic and Evolutionary Computation Conference (GECCO)},
  2024.

\bibitem{sohl2015deep}
Jascha Sohl-Dickstein, Eric Weiss, Niru Maheswaranathan, and Surya Ganguli.
\newblock Deep unsupervised learning using nonequilibrium thermodynamics.
\newblock In {\em International conference on machine learning}, pages
  2256--2265. PMLR, 2015.

\bibitem{pyribs_diffusion}
Li~Ding, Jenny Zhang, Jeff Clune, Lee Spector, and Joel Lehman.
\newblock Incorporating human feedback into quality diversity for diversified
  text-to-image generation.
\newblock {\em pyribs.org}, 2024.

\bibitem{tjanaka2023pyribs}
Bryon Tjanaka, Matthew~C Fontaine, David~H Lee, Yulun Zhang, Nivedit~Reddy
  Balam, Nathaniel Dennler, Sujay~S Garlanka, Nikitas~Dimitri Klapsis, and
  Stefanos Nikolaidis.
\newblock pyribs: A bare-bones python library for quality diversity
  optimization.
\newblock In {\em Proceedings of the Genetic and Evolutionary Computation
  Conference}, pages 220--229, 2023.

\bibitem{nikolaidis2017mathematical}
Stefanos Nikolaidis, Jodi Forlizzi, David Hsu, Julie Shah, and Siddhartha
  Srinivasa.
\newblock Mathematical models of adaptation in human-robot collaboration.
\newblock {\em arXiv preprint arXiv:1707.02586}, 2017.

\bibitem{fontaine2021iquality}
Matthew~C Fontaine and Stefanos Nikolaidis.
\newblock A quality diversity approach to automatically generating human-robot
  interaction scenarios in shared autonomy.
\newblock In {\em Robotics Science and Systems (RSS)}, 2021.

\bibitem{samvelyan2024rainbow}
Mikayel Samvelyan, Sharath~Chandra Raparthy, Andrei Lupu, Eric Hambro, Aram~H
  Markosyan, Manish Bhatt, Yuning Mao, Minqi Jiang, Jack Parker-Holder, Jakob
  Foerster, et~al.
\newblock Rainbow teaming: Open-ended generation of diverse adversarial
  prompts.
\newblock {\em arXiv preprint arXiv:2402.16822}, 2024.

\bibitem{huang2024cadre}
Peide Huang, Wenhao Ding, Jonathan Francis, Bingqing Chen, and Ding Zhao.
\newblock Cadre: Controllable and diverse generation of safety-critical driving
  scenarios using real-world trajectories.
\newblock {\em arXiv preprint arXiv:2403.13208}, 2024.

\end{thebibliography}
\end{document}